\documentclass[10pt,twocolumn,letterpaper]{article}

\usepackage{cvpr}              % To produce the CAMERA-READY version
\usepackage[dvipsnames]{xcolor}

\definecolor{cvprblue}{rgb}{0.21,0.49,0.84}
\definecolor{BUred}{rgb}{0.8, 0.0, 0.0}

\usepackage[pagebackref,breaklinks,colorlinks,citecolor=cvprblue,linkcolor=BUred]{hyperref}
\usepackage{comment}
\usepackage{cancel}
\usepackage{float}
\usepackage{placeins}
\usepackage{multirow}
\usepackage[utf8]{inputenc}
\usepackage{algpseudocode}
\usepackage{algorithm}
\usepackage{amsmath}

\def\paperID{58} % *** Enter the Paper ID here
\def\confName{CVPR}
\def\confYear{2024}

\title{GenVideo: One-shot target-image and shape aware video editing using T2I diffusion models}

\author{Sai Sree Harsha \\
Adobe\\
{\tt\small ssree@adobe.com}
\and
Ambareesh Revanur\\
Adobe\\
{\tt\small arevanur@adobe.com}
\and
Dhwanit Agarwal\\
Adobe\\
{\tt\small dhagarwa@adobe.com}
\and
Shradha Agrawal\\
Adobe\\
{\tt\small shradagr@adobe.com}
}

\begin{document}

%\vspace{-8mm}
%\begin{comment}
%%%% Figure: Teaser
\twocolumn[{
\renewcommand\twocolumn[1][]{#1}
\maketitle
\begin{center}
% \vspace{3mm}
\vspace{-2mm}
  \includegraphics[width=1\linewidth]{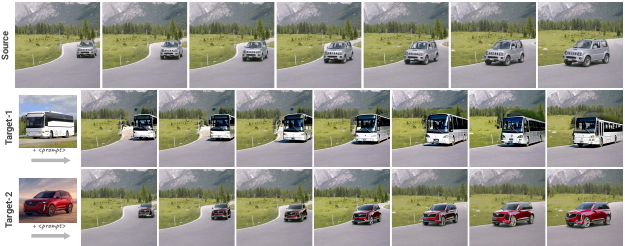}
  \captionsetup{type=figure,font=small}
  \vspace{-15pt}
  \caption{Editing the source car video based on a \textbf{target-image} and a text prompt \textit{Driving a ``white bus"/``red cadillac" down a mountain road} with GenVideo method. GenVideo can accurately replace the object in the source video, even with a target object of a different shape.}
  % \vspace{2pt}
  \label{fig:teaser}
\end{center}
}]
%\end{comment}
\maketitle
\begin{abstract}

\vspace{-5.5mm}

Video editing methods based on diffusion models that rely solely on a text prompt for the edit are hindered by the limited expressive power of text prompts. Thus, incorporating a reference target image as a visual guide becomes desirable for precise control over edit. Also, most existing methods struggle to accurately edit a video when the shape and size of the object in the target image differ from the source object. To address these challenges, we propose ``GenVideo" for editing videos leveraging target-image aware T2I models. Our approach handles edits with target objects of varying shapes and sizes while maintaining the temporal consistency of the edit using our novel target and shape aware InvEdit masks. Further, we propose a novel target-image aware latent noise correction strategy during inference to improve the temporal consistency of the edits. Experimental analyses indicate that GenVideo can effectively handle edits with objects of varying shapes, where existing approaches fail.

\end{abstract}    

\vspace{-4mm}

\section{Introduction}
\label{sec:intro}

\vspace{-2mm}

Image and video generation using diffusion models \cite{ho2020denoising,esser2023structure,dhariwal2021diffusion,nichol2021improved,nichol2021glide,podell2023sdxl,kumari2023ablating} has gained immense popularity in recent times. 
Recently, many methods have been proposed for editing visual content like images \cite{rombach2022highresolution, ramesh2022hierarchical, saharia2022photorealistic}, and videos \cite{singer2022makeavideo, zhou2022magicvideo, wu2022tuneavideo, Chai_2023_ICCV} using text prompts. 
Although several works have showcased high-quality results for the task of image editing \cite{avrahami2022blended, couairon2022diffedit, hertz2022prompt, tumanyan2022plugandplay, parmar2023zeroshot}, the utilization of diffusion models for the task of video editing has been limited. Works such as \cite{molad2023dreamix, esser2023structure}, use text-to-video (T2V) diffusion models for video editing. However, these methods have limited utility since they demand a large computational infrastructure for training along with large and diverse video dataset. 
Due to these limitations, more recent methods have started exploring inflated text-to-image (T2I) diffusion models as an alternative to T2V models for video editing in a one-shot or zero-shot fashion \cite{wu2022tuneavideo, qi2023fatezero, zhao2023makeaprotagonist, liu2023videop2p, Chai_2023_ICCV, geyer2023tokenflow, bar-tal2022text2live, ceylan2023pix2video}. Nonetheless, with the exception of Make-A-Protagonist \cite{zhao2023makeaprotagonist} which can be additionally conditioned on a target-image, existing and concurrent video editing methods that are based on T2I diffusion models focus exclusively on text-driven video editing. Therefore, these methods are not suitable for scenarios where the nature of the edit cannot be accurately expressed using only text. In other words, they are not target-image aware. We emphasize here that target-image awareness provides a precise visual guide for the desired edit. This approach enables exact replication of shapes, sizes, and textures, essential for creating content that serves use cases like creating content that aligns with a particular product specifics (like a particular model of a car). Further, previous approaches lack shape awareness and therefore, fail to make edits where the shape and size of the target-object is substantially different from the source video.

In this paper, we propose a novel target-image and shape aware \textit{InvEdit} method to effectively identify a region-of-interest in the video for a given text prompt and the target-image. This is done using the underlying T2I diffusion model similar to \cite{couairon2022diffedit} but for videos and using additional image conditioning. The strength of our method lies in InvEdit and its simplicity with which it can be adapted to any image conditioned diffusion model to get target-image and shape aware mask for video editing. Localized mask guidance is often crucial for precise edits and avoiding leakage of edits to the unmasked region but obtaining target-image and shape aware masks is a challenging problem due to which previous methods are not able to handle edits of substantially different shapes and sizes.

We further address the key challenge of maintaining the temporal consistency in the edited video. Some of the previous works in this research area fail to effectively address this challenge because of their reliance on source-based Neural Layer Atlases (NLA) \cite{Chai_2023_ICCV} or source-based inter-frame feature propagation \cite{geyer2023tokenflow}. Hence, the temporal consistency of the video generated by these methods is primarily dictated by the source video. 
Naturally, these methods fail when the target-object is substantially different from the source. To address these concerns, we introduce a novel target-image aware \emph{latent correction} strategy to blend the inter-frame latents of the diffusion model on the fly during inference to improve the inter-frame temporal consistency of the target-object in the edited video. Our approach uses \emph{InvEdit} mask guidance to enable temporal consistency even when the target-object has a different shape and size than the object in the source video. We show extensive quantitative and qualitative comparisons with state-of-the-art video editing methods and conclude that our approach outperforms previous and concurrent approaches on a diverse set of videos. Our contributions can be summarized as follows:

\begin{itemize}
    \item We introduce \emph{GenVideo}, the first pipeline for target-image and shape aware editing of videos using image diffusion models that can make temporally consistent edits for target objects of varied shapes and structures.

    \item We introduce \emph{InvEdit}, a novel target-image and shape aware, zero-shot approach for generating video masks using image diffusion models. This mask enables us to identify the region of interest for editing objects of varying shapes and sizes. 
    
    \item We introduce a target-image and shape aware \emph{latent correction} strategy to enforce inter-frame temporal consistency in the edited video, where existing approaches fail.

\end{itemize}

\section{Related Work}
\label{sec:related}

\textbf{Text Driven Image and Video Editing:}  
Before the advent of the diffusion models \cite{song2023denoising, ho2020denoising}, generative adversarial networks (GANs) have been used to edit images ~\cite{Oord2017, Yu2022, Kang2023, Patashnik2021, Wang2018, revanur2023coralstyleclip}. Recently, generative diffusion  models ~\cite{ramesh2022hierarchical, rombach2022highresolution, saharia2022photorealistic} trained on internet scale image and text data have achieved significantly higher quality and diversity of image generation than GANs. Inspired by the success of these T2I diffusion models, text-to-video diffusion models ~\cite{ho2022imagen, singer2022makeavideo, hong2022cogvideo, luo2023videofusion} have been developed for video generation. Authors in ~\cite{singer2022makeavideo} inflate the pretrained T2I diffusion models with additional temporal attention layers to generate videos, while the recent works like ~\cite{ho2022imagen} use a 3D-UNet for video generation.  Along with generation, editing of existing images and videos has also received significant interest. SDEdit \cite{meng2021sdedit} carefully adds noise to an input image and then denoises them using diffusion models to achieve the desired edits. Prompt-to-prompt (P2P) \cite{hertz2022prompt} allows text guided editing of images by controlling the cross-attention maps.  In the realm of video editing, Tune-A-Video (TAV) \cite{wu2022tuneavideo}  finetunes the inflated T2I model on an individual source video to generate edited videos with similar motion. Video-P2P \cite{liu2023videop2p} and FateZero \cite{qi2023fatezero} achieve controllable edits in videos by altering the cross-attention maps. Pix2Video \cite{ceylan2023pix2video} achieves video editing by altering the self-attention features during inference. Text2Live \cite{bar-tal2022text2live} utilizes NLA and then edits each layer independently using text guidance. Text2Video-Zero \cite{khachatryan2023text2video} uses inflated self-attention layers along with ControlNet \cite{zhang2023adding} to do text guided video editing. Nonetheless, these works using T2I diffusion models exclusively focus on image and video editing via text, and cannot be used for edits which are indescribable by text alone. Hence, they are not target-image aware which is often a crucial aspect for accurate video editing \cite{zhao2023makeaprotagonist}.  

\vspace{1mm}

\noindent \textbf{Target-Image Aware Visual Content Editing:} To cater to visual content edits  that isn't easily or accurately described by text alone, models like DALLE--2 \cite{ramesh2022hierarchical} and Gen--1 \cite{esser2023structure} have been developed to make target-image aware edits to images and videos by leveraging the CLIP \cite{radford2021learning} image embedding. While DALLE--2 edits images, Gen--1 can edit videos but the drawback here is that the entire image, including the background, influences the edits due to the 
the lack of mask guidance. 
Subject driven solutions like Dreamix ~\cite{molad2023dreamix}, which can edit videos conditioned on the target-image data, typically require multiple target images for learning the concept/target object accurately and also require large video diffusion models as used in \cite{ho2022imagen}. While they can generate and edit videos based on a single image, the edited videos fail to maintain the fidelity of the edits to the target-object in the target-image. Make-A-Protagonist \cite{zhao2023makeaprotagonist} addresses these limitations by using an image guided T2I diffusion model called SD-unCLIP \cite{ramesh2022hierarchical}. Their pipeline can achieve target-image aware video object editing using a single target-image and employ mask-guided inference to do localized edits. However, the edits still suffers from lack of temporal consistency and the approach fails when the target object differs in shape and size from the source object present in the source video. This is because the mask used for editing the video is generated only using the source video and has no awareness of the target-object.

\vspace{1mm}
\noindent \textbf{Video Editing with Temporal Consistency:} Aforementioned methods suffer from lack of inter-frame temporal consistency in the edited video. A concurrent work, \emph{TokenFlow} \cite{geyer2023tokenflow} improves the temporal consistency by inter-frame propagation of the self-attention features using a nearest neighbor field in the feature space of source video. However, both the editing and temporal consistency in \emph{TokenFlow} are not target-image and shape aware. Thus, it cannot make edits where the target-object differs a lot in shape and size to the source object. Moreover, it does not offer a way to localize the edits due to the lack of mask guidance and also requires manually choosing key frames for imposing temporal consistency. 
\emph{StableVideo} \cite{Chai_2023_ICCV} uses NLA of the source video to enforce temporal consistency and hence, also lacks target-image and shape awareness for accurate shape changing edits. Another work \cite{lee2023textvideoedit}  addresses shape-aware editing by formulating a deformation map which maps the original atlas into edited atlas. While the method achieves favorable results, NLA mapping takes tens of hours of training and an inaccurate NLA mapping results in objects with missing parts and substantial undesirable background changes. Our work \emph{GenVideo} addresses these limitations by providing the first ``target-image and shape aware" video editing using T2I models with only \emph{30 mins of training and 2 mins of inference} using localized mask guidance to avoid background changes.

\begin{figure}[t!]
    \centering
    \includegraphics[width=\linewidth]{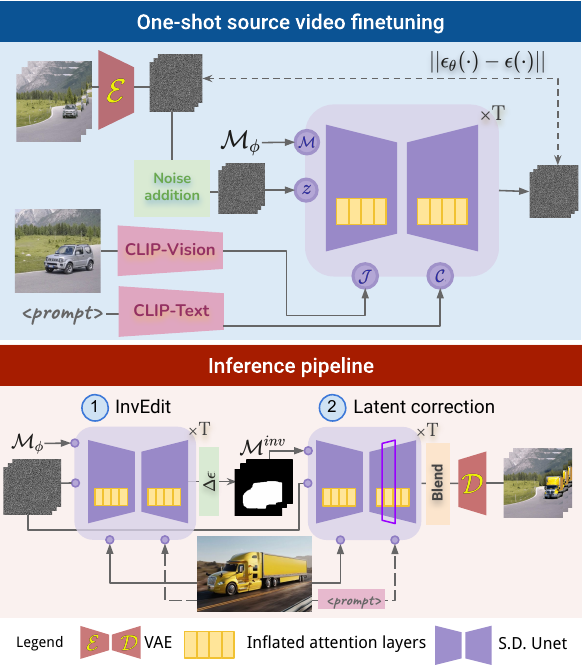}
    \vspace{-6mm}
    \caption{Overview of \textit{GenVideo}.  Inflated attention layers are finetuned during source video finetuning. During inference, \emph{InvEdit} predicts a  region to edit and \emph{latent correction} uses that mask to improve the inter-frame temporal consistency. $\mathcal{M}_{\phi}$ - ``no mask". }
    \label{fig:method}
    \vspace{-6mm} 
\end{figure}

\section{Approach}
\label{sec:method}

\textit{GenVideo} aims to edit the given source video based on a target text prompt and a target-image containing an object of any shape while maintaining  temporal consistency.  
More formally, given an input source video 
$\mathcal{V}^{src} = [I^{src}_{1}, \cdots, I^{src}_{N}]$ composed of $N$ frames containing a source object, a source text prompt $\mathcal{P}^{src}$ describing the source video, a target-image $I^{trg}$ containing the target-object and a target text prompt $\mathcal{P}^{trg}$ describing the desired edit to the source video, \textit{GenVideo} generates a target video  $\mathcal{V}^{trg} = [I^{trg}_{1}, \cdots, I^{trg}_{N}]$ which preserves the motion of the input source video but replaces the source object with a new target-object from the target-image. 
The entire training and inference pipeline is summarized in Fig. \ref{fig:method}. First, we fine-tune an inflated Stable Diffusion unCLIP (SD-unCLIP) model \cite{ramesh2022hierarchical, zhao2023makeaprotagonist} on the source video using the reconstruction loss from standard LDMs \cite{rombach2022highresolution} (Sec. \ref{sec:finetune}). Then, we employ our novel target-image and shape aware mask generation approach, called \textit{InvEdit}, where we use the fine-tuned model to infer a region-of-interest where the edits need to be localized (Sec. \ref{sec:invedit}). Finally, we introduce a novel \emph{latent correction} approach to improve the inter-frame temporal consistency (Sec. \ref{sec:latentnoise}).

\vspace{1mm}

\begin{figure*}[ht!]
    \centering
    \includegraphics[width=\linewidth]{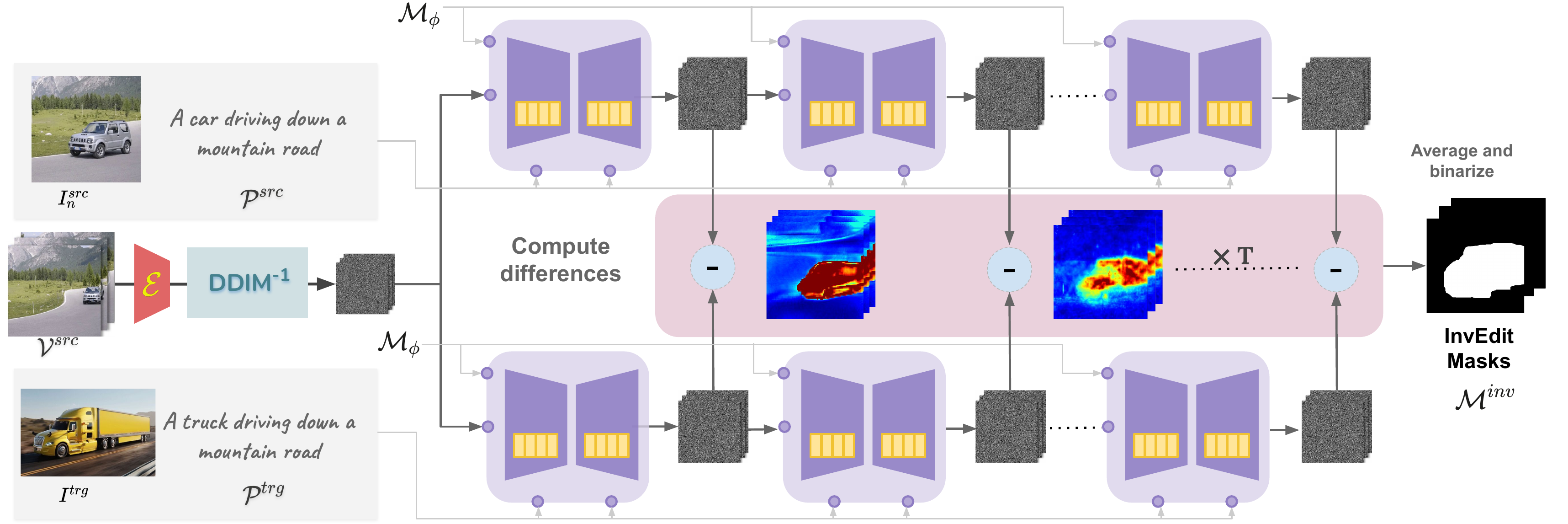}
    \vspace{-6mm}
    \caption{\textit{InvEdit} approach - the mask is generated by first iteratively computing noise differences across multiple timesteps for the source denoising branch and target denoising branch. Then, these differences are averaged and binarized to obtain the \textit{InvEdit} mask.}
    \label{fig:invedit}
    \vspace{-6mm}
\end{figure*}

\begin{figure}
    \centering
    \includegraphics[width=\linewidth]{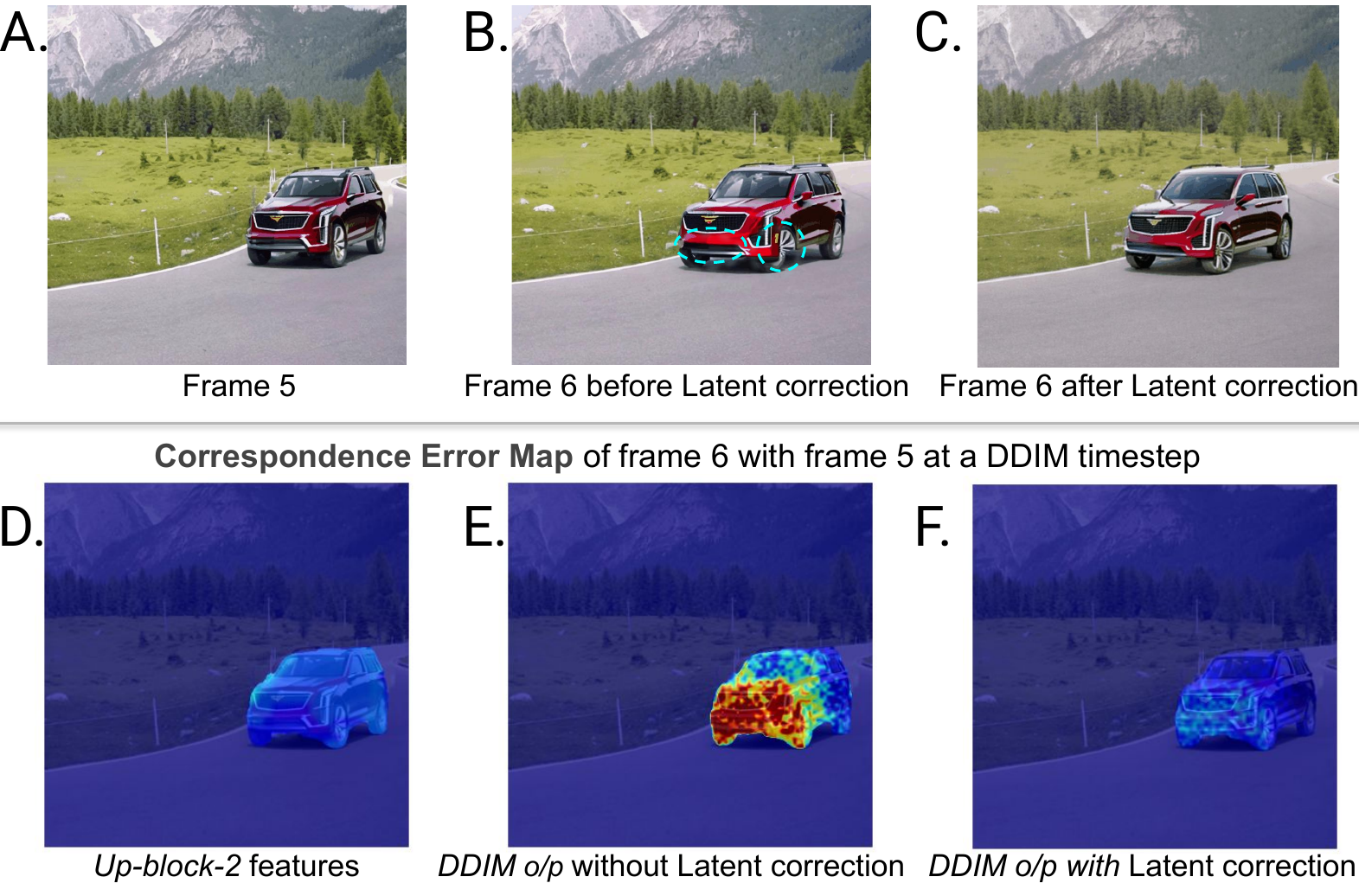}
    \vspace{-7mm}
    \caption{\emph{Latent correction} strategy corrects the latent noise using UNet features from the previous and successive video frame. Correspondence Error (CE) map computed using the ground truth (source video correspondences) shows that the DDIM o/p of the model before correction has high CE (\textbf{E.}) which our \textit{latent correction} strategy fixes (\textbf{F.}) using \texttt{Up-block-2} feature correspondences which have low CE (\textbf{D.}).}
    \label{fig:latent_correction}
    \vspace{-5mm}
\end{figure}

\subsection{Fine-tuning on the source video}
\label{sec:finetune}
\vspace{-2mm}

The finetuning procedure of the inflated pretrained T2I diffusion model is in line with the TAV method \cite{wu2022tuneavideo} (Fig. \ref{fig:method}). Unlike \cite{wu2022tuneavideo}, we use the SD-unCLIP model \cite{ramesh2022hierarchical} which conditions the generation on both target-image and the text prompt
 \cite{zhao2023makeaprotagonist}. This model uses CLIP-vision branch to obtain an image embedding $\mathcal{J}^{*}$ for an input reference image and uses the CLIP-text branch to obtain the text embedding $\mathcal{C}^{*}$.  
As part of the inflation process, spatial self-attention is inflated into spatio-temporal attention (ST-attn), and an additional temporal attention (T-attn) is introduced after the ST-attn and cross-attention blocks. See Suppl. for details.

\subsection{InvEdit mask generation}
\label{sec:invedit}
\vspace{-2mm}
In this section, we describe the details of \textit{InvEdit} -- our novel zero shot, target-image and shape aware mask generation strategy using the finetuned diffusion model from Sec. \ref{sec:finetune}. Existing methods\cite{zhao2023makeaprotagonist, qi2023fatezero} compute masks using the source video alone and hence have no shape awareness regarding the target object's relative shape and size (for example, changing a car to a bus). InvEdit is an adaptation of DiffEdit \cite{couairon2022diffedit} for videos and includes target-image and shape awareness.

\noindent \textbf{\emph{InvEdit} steps:} First, we perform the DDIM inversion \cite{song2023denoising} to transform the source video into the corresponding random latent noise $\mathcal{Z}^{src}_T = [z^{src}_{T, 1}, z^{src}_{T, 2}, \cdots, z^{src}_{T, N}]$. Then, we denoise the latent code $\mathcal{Z}^{src}_{T}$ using the fine-tuned inflated SD-unCLIP model from Sec. \ref{sec:finetune} using deterministic DDIM sampling for both the source and target branches as shown in Fig. \ref{fig:invedit}. For the source branch, we denoise $\mathcal{Z}^{src}_{T}$ using the source text prompt $\mathcal{P}^{src}$ and a randomly chosen frame $I^{src}_n$ from $\mathcal{V}^{src}$ as conditioning inputs for the fine-tuned SD-unCLIP model. Similarly, in parallel, we repeat DDIM sampling using the target text prompt $\mathcal{P}^{trg}$ and the foreground object in the target-image $I^{trg}$ as conditioning inputs. We segment the foreground object from the target-image using GroundedSAM \cite{kirillov2023segany, liu2023grounding}. 

We compute the difference of the predicted noise by the denoising UNet model (denoted by $\varepsilon_\theta$) obtained at each denoising timestep in the source and target branches. More formally, for every $I^{src}_n$ in $\mathcal{V}^{src}$, we compute $\Delta\varepsilon_{t, n} = \mathop{abs}(\varepsilon^{src}_{t, n} - \varepsilon^{trg}_{t, n})$, where $\textit{ }\varepsilon^{u}_{t, n} = \varepsilon_\theta\left({z}^{u}_{t, n}, t , \mathcal{C}^{u}, \mathcal{J}^{u}_n\right), u \in \{src, trg\}$. At each DDIM denoising step we obtain,
\vspace{-1mm}
\begin{equation}
\begin{aligned}
    \mathcal{Z}^{src}_t = \mathrm{DDIM}(\mathcal{Z}^{src}_{t+1}, \epsilon^{src}_{t+1}) ,
    \mathcal{Z}^{trg}_t = \mathrm{DDIM}(\mathcal{Z}^{trg}_{t+1}, \epsilon^{trg}_{t+1}), \\
\end{aligned}
\end{equation}
\vspace{-1mm}

where $t \in \{ T-1, \cdots, 1\}$  denotes the time step. 

\vspace{0.5mm}
These differences $\Delta\varepsilon_{t, n}$, represented as heat maps in Fig. \ref{fig:invedit}, are averaged over multiple denoising time steps and binarized to get the target-aware \textit{InvEdit} mask for each frame of the video. In Fig. \ref{fig:invedit}, the \textit{InvEdit} mask is able to determine that edits are to be placed in a region shaped like a truck rather than the car, as the truck is much bigger than the car. We denote the masks for $N$ frames by $\mathcal{M}^{inv} = [M_{1}, M_{2}, \cdots, M_{N}]$, 
where $M_n=binarize(\mathop{mean}_{t \in [t_i, t_j]}(\Delta\varepsilon_{t, n}))$. The \textit{InvEdit} mask is used to identify the region in which 
the target-image embeddings are injected into the features of the ResNet blocks of the UNet along with the timestep embeddings \cite{ramesh2022hierarchical}.

\noindent\textbf{\emph{InvEdit} intuition:} Our intuition follows DiffEdit \cite{couairon2022diffedit} for text driven image editing. We extend it for our target-image aware video editing use case. During DDIM denoising, the SD-unCLIP model will yield different noise estimates given different text and image conditioning. The noise estimates will be different in regions where the latent code will eventually decode different shapes, colors and textures depending on the conditioning. For the background, on the other hand, there is little change in the noise estimates. The difference between the noise estimates can thus be used to infer a mask that identifies the parts of each video frame need to be edited. 

\subsection{Latent noise correction via self-consistency}
\label{sec:latentnoise}
\vspace{-2mm}

While \textit{InvEdit} mask is able to accurately identify the regions to edit, it does not address the temporal consistency of the object within the region across generated frames. As an example, consider the edit where the ``\textit{silver car}" is edited to a ``\textit{red cadillac}" shown in Fig. \ref{fig:latent_correction}. While the shape of the car generated by \textit{InvEdit} mask in Frame 6 (see Fig. \ref{fig:latent_correction}{\color{BUred} B}) appears similar to Frame 5 (see Fig. \ref{fig:latent_correction}{\color{BUred}A}), it has a different stylistic appearance in the front and in the side. One trivial way to address this is by computing the optical flow of the source object in the video and then imposing this flow on the latent noise features $\mathcal{Z}$. However, this solution will struggle considerably even in typical situations where the target object is of a different shape. Thus, the issue of temporal inconsistency becomes challenging given the model has not seen the target-image before. We address it by introducing a \textit{latent correction} strategy during inference. This correction is a blending strategy in the latent $\mathcal{Z}$ space to improve the inter-frame temporal consistency of the edited video. This is a three-step process (also see Supplementary Material (SM)): 

\begin{enumerate}
    \item \textbf{Inter-frame latent field computation:} During each denoising time step $t$ of inference, 
    we utilize the features of the \texttt{Up-Block-2} of the UNet denoted as $[f^{t}_{1}, \cdots, f^{t}_{N}]$ for estimating latent correspondence maps/field using nearest neighbors between these features of the consecutive frames. First, we compute the nearest neighbors field $\mathcal{N}_{i \pm}(\cdot)$ defined as $ 
        \mathcal{N}^{t}_{i \pm}[p] := \mathrm{arg max}_{q} {d} (f^{t}_{i}[p], f^{t}_{i \pm 1}[q]). \label{eq:nnfield} $ This field is the mapping from the spatial locations $p$ in the features of $i^{th}$ frame to its nearest neighbor $q$ (in terms of cosine similarity ${d}$) in the features of the $(i\pm 1)^{th}$ frame. This field is the mapping from the spatial locations $p$ in the features of $i^{th}$ frame to its nearest neighbor $q$ in the features of the $(i\pm 1)^{th}$ frame.

    \vspace{1mm}
    \item \textbf{Blending using inter-frame latent field:}  Starting from the DDIM inversion of the source video $\mathcal{Z}^{src}_T$, at each denoising timestep $t$ during inference, we blend the latents $\mathcal{Z}_t = [z_{t, 1}, z_{t, 2}, \cdots, z_{t, N}]$  of adjacent frames inside the \textit{InvEdit} mask region in the latent space of the decoder $\mathcal{D}$ of the VAE of SD-unCLIP. 
    The blended latents $\Tilde{\mathcal{Z}}_{t} = [\Tilde{z}_{t, 1}, \cdots, \Tilde{z}_{t, N}]$ at timestep $t$ are given by,   
    \vspace{-2mm}
    \begin{align*}
        \Tilde{z}_{t, i}[p] &= w_{-1} (M_{i} \odot  z_{t, (i-1)}[\mathcal{\hat{N}}^{t}_{i-} (p)])  + w_{0}( M_{i} \odot z_{t, i}) \\
        &\quad + w_{1}( M_{i} \odot z_{t, (i+1)}[\mathcal{\hat{N}}^{t}_{i+} (p)]) \\
        &\quad + (1 - M_{i}) \odot z_{t, i},
    \end{align*}
    \vspace{0mm}
where $w_{-1}, w_{0}, w_{+1}$ are non-negative weight hyperparameters which add up to 1 and $\mathcal{\hat{N}}^{t}_{i \pm}[p]$ is the nearest neighbors field above upsampled to match the dimension of the $z_{t}$. This blending happens at each inference timestep $t$ for $t\geq T-5$. 

    \item \textbf{Background preservation:} We also correct the latent noise corresponding to background regions using the inverse \textit{InvEdit} mask, \emph{i.e.,} $(1 - M_{i})$ and denoise only the masked area \cite{song2023objectstitch}. 
    This is achieved by retaining the clean latent corresponding to the source video frame outside the masked region as given by $\Tilde{z}_{t, i} =  (1 - M_{i}) \odot \mathcal{E}({I}^{src}_{i}) +  (M_{i}) \odot  \Tilde{z}_{t, i} $, where $\mathcal{E}$ is the encoder of VAE. We skip this step when the background of the target video is expected to be different from the source.

\end{enumerate}

\vspace{1mm}
\noindent \textbf{Why Up-block-2?} We use the features of \texttt{Up-block-2} since it demonstrates a low correspondence error (CE) than the CE in the latent noise obtained after DDIM step. Continuing the example of the car in Fig. \ref{fig:latent_correction}, we first compute the feature correspondences in consecutive frames using RAFT optical flow \cite{teed2020raft} on the source video frames. This serves as a ground truth correspondence for this example since the edited object has the same shape as the source object. Then we compute the correspondences of the features of \texttt{Up-block-2} in consecutive frames and find that these features have a low CE rate as shown by the heatmap in 
Fig. \ref{fig:latent_correction}{\color{BUred}D}. On the other hand, the CE rate is higher for latent noise computed after the DDIM Step 
(Fig. \ref{fig:latent_correction}{\color{BUred}E}). With our \textit{latent correction} strategy, the CE is lowered since the proposed blending strategy improves the consistency of the latent noise features across consecutive frames as demonstrated by 
Fig. \ref{fig:latent_correction}{\color{BUred}C} and Fig. \ref{fig:latent_correction}{\color{BUred}F}. We show in our experiments that this process improves the temporal consistency of the edited target video.

\begin{figure*}[t!]
    \includegraphics[width=\textwidth]{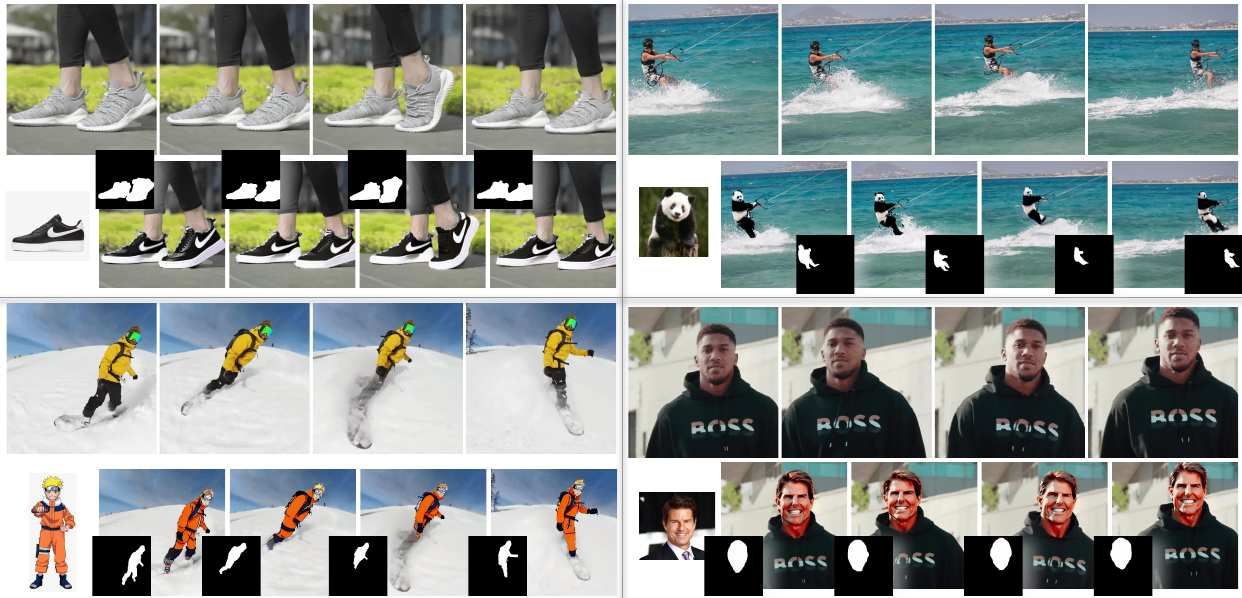}
    \vspace{-6mm}
    \caption{Source videos, Target images, \textit{InvEdit} mask and the \textit{GenVideo} results. Top: \textit{``A person wearing a \cancel{gray} black shoe."}, and \textit{``A \cancel{man} panda rides a kite surfboard in deep waters."} Bottom: \textit{``\cancel{man} naruto skiing on snow"}, and \textit{`` \cancel{man} Tom Cruise walking down the street"}. %We recommend zooming in for best viewing.
    }
    \label{fig:results}
    \vspace{-4mm}
\end{figure*}

\vspace{-2mm}
\section{Experiments}
\label{sec:exp}

\vspace{-1mm}
\subsection{Implementation details}
\vspace{-2mm}
For all of our experiments, we use the SD-unCLIP version of Stable Diffusion v2.1 T2I model \cite{ramesh2022hierarchical}. We inflate the model with temporal layers as in \cite{wu2022tuneavideo}. See SM for more details on model inflation. During source finetuning, we train the inflated layers using $16$ source video frames with a learning rate of $1\times10^{-5}$ for $400$ iterations. We use DDIM sampler during inference to obtain the \textit{InvEdit} mask on the DDIM-inverted source video by computing the mean of $\Delta \epsilon_t$ across $t\in[0.8\times T, T]$ where $T=50$ and use a difference threshold of 0.6 to binarize the heatmap. During the \emph{latent correction},  we restrict the nearest neighbour search to a $4\times4$ window of pixels and set $w_{-1}=0.1$, $w_{0}=0.8$ and $w_{1}=0.1$ across the experiments. We also fix classifier-free guidance scale to $12.5$ during training and inference.

\begin{figure}[t!]
    \centering
    \includegraphics[width=\linewidth]{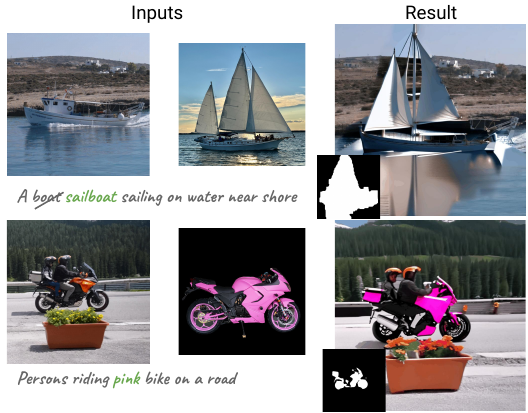}
    \vspace{-6mm}
    \caption{Results of \textit{InvEdit} on zero-shot image editing showing its capability for target-objects of varying shape and size.}
    \vspace{-6mm}
    \label{fig:image_editing}
\end{figure}

\subsection{Applications}
\vspace{-2mm}

    \textbf{Shape-aware video object editing :} Our approach can be used to do target-image and shape aware video object editing. Fig. \ref{fig:results}, along with the second series in Fig. \ref{fig:teaser}, showcase the outcomes of temporally consistent video object editing using \textit{GenVideo}. This task is difficult as the mask needs to be target-image and shape aware to effectively identify the region of interest for localized edits. For instance, in the \emph{man} $\longrightarrow$ \emph{naruto}, \emph{man} $\longrightarrow$ \emph{panda} and \emph{silver car} $\longrightarrow$ \emph{white bus} examples, \textit{InvEdit} mask is able to accurately identify the mask for \emph{naruto}, \emph{panda}, and the \emph{white bus} respectively. Furthermore, our \emph{latent correction} strategy is able to correct the UNet latents on the fly during inference. 
    
    \vspace{-0.7mm}
    
     \noindent \textbf{Zero-shot image editing:} Moreover, our approach is also capable of zero-shot image editing for objects of varying shapes and sizes.  This can be observed in Fig. \ref{fig:image_editing}, which demonstrates object shape changes (like \emph{boat} $\longrightarrow$ \emph{sailboat}) in images. Additional examples and results are provided in the SM we have made available.

\begin{table}[t]
\centering

\caption{Quantitative comparison of the state-of-the-art video editing methods. For the user study, \emph{Text}, \emph{Image}, and \emph{Visual} refer to the average rank in target text alignment, target-image alignment, and visual quality of the edited video respectively.  \label{tab:comp} 
}
\vspace{-2mm}
\setlength\tabcolsep{2pt}
\resizebox{\linewidth}{!}{
    \begin{tabular}{l c c c | c c c}
    \toprule
    \multirow{2}{*}{Method} & \multicolumn{3}{c}{Model scoring metrics $(\uparrow)$} & \multicolumn{3}{c}{User Study $(\downarrow)$} \\
    
    \cmidrule(l{1mm}r{1mm}){2-4}
    \cmidrule(l{1mm}r{1mm}){5-7}
    
     & $\mathrm{CLIP}$-$\mathrm{T}$ & $ \mathrm{DINO}$ & $\mathrm{Temp}$ & Text& Image& Visual \\ \midrule
    TAV \cite{wu2022tuneavideo} & 0.238 & 0.236 & 0.957 & 3.6 & 3.3 & 4.2\\ 
    StableVideo \cite{Chai_2023_ICCV} &  0.234 & 0.189 & 0.980 & 4.3 & 4.3 & 3.7 \\
    TokenFlow \cite{geyer2023tokenflow} & 0.231 &  0.216 & \textbf{0.985} & 3.3 & 3.8 & \textbf{2.1} \\
    FateZero \cite{qi2023fatezero} &  0.235 & 0.262 & 0.951 & 3.9 & 3.6 & 3.4 \\
    Make-a-Pro \cite{zhao2023makeaprotagonist} & 0.234 & 0.195 & 0.949 & 4.0 & 4.1 & 5.0 \\ 
    Ours \emph{GenVideo} & \textbf{0.241} & \textbf{0.374} &  0.967 & \textbf{1.7} & \textbf{1.8} & 2.3 \\ 
    \bottomrule
    \end{tabular}
}
\vspace{-5mm}

\end{table}

\vspace{-1mm}
\subsection{Comparison with baselines}
\vspace{-2mm}
\begin{figure*}
     \includegraphics[width=\linewidth]{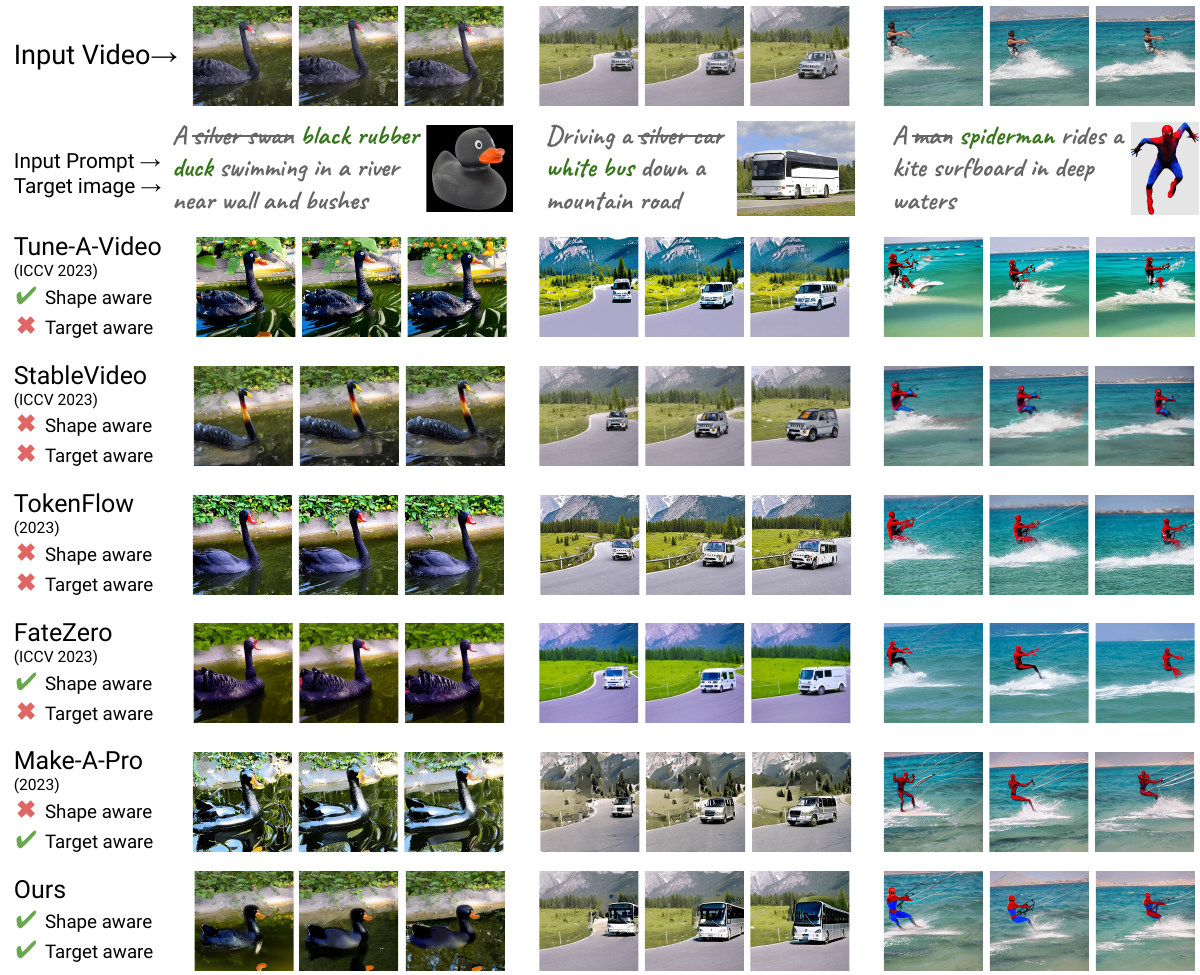}
     \vspace{-7mm}
    \caption{Qualitative comparison of \textit{GenVideo} against SOTA. Shape aware methods demonstrate the ability to edit the shape of object. Target-aware methods condition the diffusion model on both target text prompt and target-image. \textit{GenVideo} combines both target awareness and shape awareness, thus allowing it to infer shape and appearance based on the target-image and target text. This offers much superior alignment to the target text and target-image. First two columns show large shape variation while the the third shows results for same shape edits. 
    }
    \label{fig:comparison}
    \vspace{-5mm}
\end{figure*}

\begin{figure}[t!]
    \includegraphics[width=0.97\linewidth]{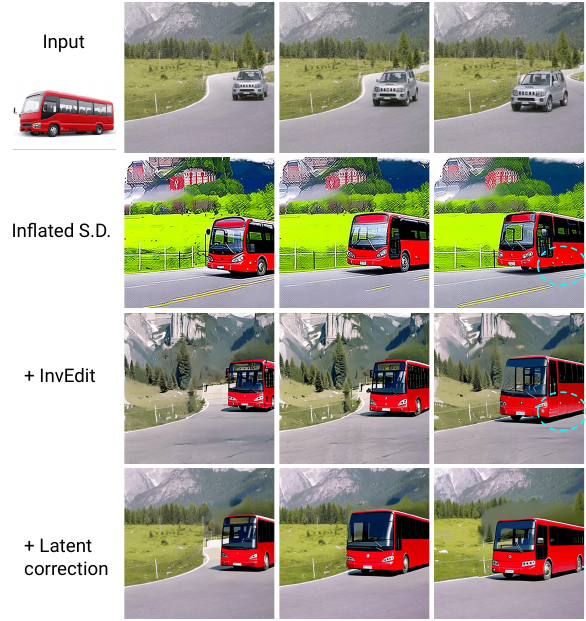}
    \vspace{-3mm}
    \caption{Ablation study of the \textit{GenVideo} method. Prompt: \textit{Driving a \cancel{silver car} red bus down a mountain road.} }
    \label{fig:ablations}
    \vspace{-8mm}
\end{figure}

We do qualitative and quantitative comparisons with the following five existing approaches: 1) Tune-A-Video (TAV) \cite{wu2022tuneavideo}, 2) StableVideo \cite{Chai_2023_ICCV}, 3) TokenFlow \cite{geyer2023tokenflow}, 4) FateZero \cite{qi2023fatezero} and 5) Make-A-Protagonist (MAP) \cite{zhao2023makeaprotagonist}. We use target reference image only with our method and \cite{zhao2023makeaprotagonist}. 

\vspace{1mm}
\noindent \textbf{Qualitative Results:}  
Fig. \ref{fig:comparison} shows a comparison of our \emph{GenVideo} approach with the above baselines. The edits in TAV\cite{wu2022tuneavideo} are not target-image aware and cannot be localized due to lack of mask. Hence, it doesn't guarantee fidelity to the target-image and also ends up modifying the background substantially. StableVideo is not target-image aware and hence, cannot generate indescribable features of the target-image. While it provides temporal consistency using NLA, these atlas are computed only using the source video and have no awareness of target-object. Hence, they fail when the target-object has a different shape and size. TokenFlow also does not provide temporal consistency for target objects of varying shapes and sizes since the nearest neighbor field they use is based solely on the source video frames. Besides both TokenFlow and StableVideo are not target-image aware and hence, do not align well with the target-image. FateZero can change shapes to some extent but since their mask is again source video-based, they cannot create large variations in the shape (like car to bus). MAP also offers source video based mask guidance and hence, cannot substantially change shapes. Our results perform temporally consistent video editing when the target-object has a substantially different shape and size while maintaining fidelity to the target-image.  

\vspace{-1.1mm}
\noindent \textbf{Quantitative Results:} Table \ref{tab:comp} shows a quantitative comparison between our approach and the state-of-the-art (SOTA) methods. We use trained CLIP \cite{radford2021learning} model to compute the average similarity of generated video frames to target text prompt which is denoted as CLIP-T in Table \ref{tab:comp}. To evaluate temporal consistency, we use CLIP model to compute average inter-frame similarity score for pairs of consecutive frames of the edited video. We denote this score as \emph{Temp} in Table \ref{tab:comp}. To measure alignment with target-image, we use trained DINO \cite{caron2021emerging} to compute the similarity between the video frames and the target-image. We also conduct a user study to evaluate the edited videos on three metrics: target text alignment, target-image alignment, and visual quality of the edited video. We asked 20 subjects to rank the edited videos for the six approaches ( \emph{i.e.,} ours and the five baselines mentioned above) on the above three metrics by asking each subject 9 sets of comparisons. The results are tabulated in Table \ref{tab:comp}. These results show our proposed approach outperforms existing approaches in both target text and target-image alignment. This is reflected in higher CLIP-T and DINO scores as well as in the user study where \emph{GenVideo} achieves the best average rank for text and image alignment. \emph{TokenFlow} and \emph{StableVideo} achieved a marginally better score in inter-frame temporal consistency (measured by \emph{Temp}) but this is because they use the source video ground truth for enforcing temporal consistency. This works well for consistency but the output videos do not align well with the target-image as reflected in much lower DINO scores. In the user study, \emph{TokenFlow} achieved the best rank for visual quality outperforming our approach by a small margin. Again, this is because of the source video's flow-based feature propagation which leads to good quality videos even though they rank lower in the text and image alignment. Our results show that our approach outperforms the other approaches in text and image alignment while being close to the state-of-the-art in temporal consistency and visual quality. Note that to be fair we do not expect to be the best in temporal consistency and quality because we are creating novel views of the target object from a single target reference image. Rather, the strength of our method lies in being able to create target-image and shape aware masks for localized video edits and enforce temporal consistency in masks in a training-free fashion while ensuring image and text alignment.

\subsection{Ablations}
\vspace{-2mm}
 We show the ablations for our \textit{InvEdit} and \textit{latent correction} strategies in Fig. \ref{fig:ablations}. The first row shows the inputs, \emph{i.e.,} target-image of a red bus and the source video of a silver car driving down a mountain road. The second row shows the edited video frames with a finetuned inflated SD-unCLIP model with target-image guidance without the \emph{InvEdit} mask. Note that without mask localization, the fidelity of the edited bus is of low quality and the background gets substantially modified. In third row, using \textit{InvEdit} mask localization, we notice that the fidelity of the bus to the target-image is improved and the background is not substantially modified. However, we notice that in the third column, the wheel of the bus disappears. Our \emph{latent correction} strategy is able to correct such inconsistencies by imposing temporal self-consistency between frames and preserving the background as shown in the fourth row. 

 \begin{figure}[t]
     \centering
     \includegraphics[width=\linewidth]{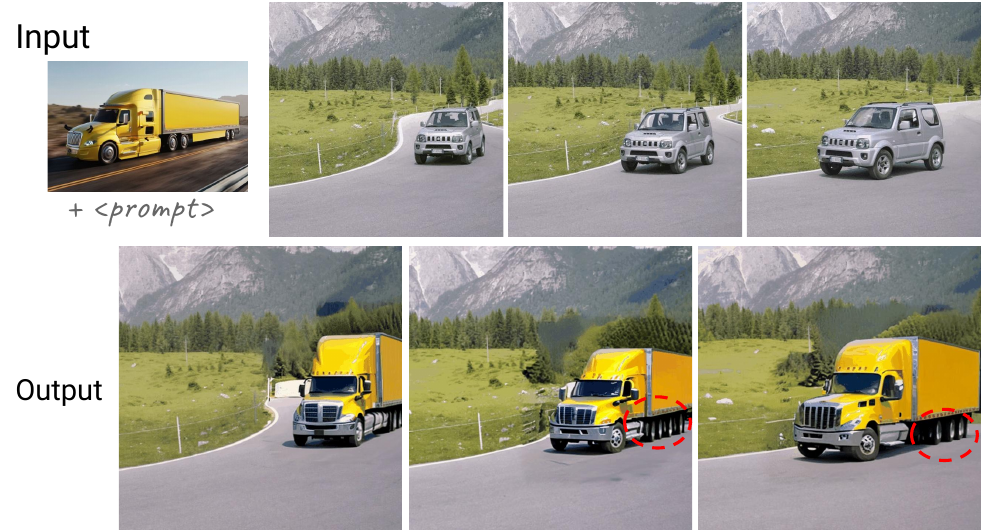}
     \vspace{-6mm}
     \caption{Limitation of \textit{GenVideo} in fine-grained editing of videos. Prompt: \textit{Driving a \cancel{silver car} yellow truck down the mountain road.}}
     \label{fig:limitation}
     \vspace{-5mm}
 \end{figure}

\section{Conclusion }
\label{sec:conc}
% \vspace{-2mm}

We introduce \textit{GenVideo}, a pipeline for target-image and shape aware video editing using image diffusion models. 
The proposed pipeline allows localized edits with target-objects using the proposed \textit{InvEdit} mask and enforces temporal consistency between the frames using \textit{latent correction} strategy. Results show that \textit{GenVideo} outperforms existing methods both qualitatively and quantitatively on the video editing task. 
\textbf{Limitations and Future Work:} The underlying SD-unCLIP model may have limitations regarding the quality and diversity of the generated content, thus affecting the edit quality. 
The \emph{latent correction} approach for inter-frame temporal consistency may not completely eliminate fine-grained inconsistencies, especially for complex objects like a truck with several wheels as shown in Fig. \ref{fig:limitation}. Furthermore, similar to previous other methods \cite{qi2023fatezero, wu2022tuneavideo}, our method cannot generate entirely new motion like changing motion from driving to flying. As part of further work, we aim to test our approach with image conditioned video diffusion models like \cite{girdhar2022emuvideo}.

\clearpage
\setcounter{page}{1}

\maketitlesupplementary

\section{Architecture details \label{sec:arch_supp}}

This section provides an overview of the underlying model and explains how the features are passed through the pipeline during training and inference.

\begin{figure}[t]
    \centering
    \includegraphics[width=\linewidth]{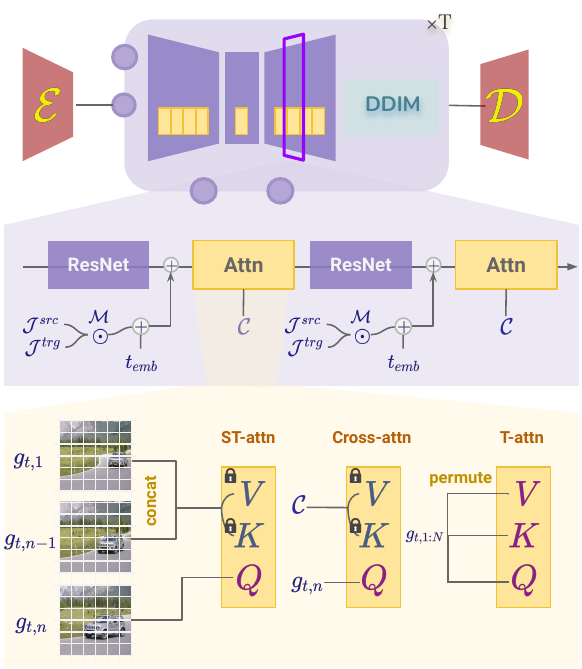}
    \caption{Architectural diagram. \textit{Top to bottom}: UNet architecture with VAE, UNet block architecture, Attention layer inflation of ST-attn, Cross-attn, and T-attn. }
    \label{fig:suppl_archi}
\end{figure}

\vspace{2mm}
\noindent \textbf{Base model}: We use the SD-unCLIP model, a fine-tuned version of the Stable Diffusion v2.1 text-to-image model that accepts CLIP image embedding and the text prompt as conditional input. The network broadly consists of the VAE autoencoder $\{\mathcal{E}, \mathcal{D}\}$, the latent denoising UNet $\varepsilon_{\theta}(\cdot)$, and CLIP conditional models (image branch and text branch) which extract an image embedding $\mathcal{J}$ and a text embedding $\mathcal{C}$. As shown in Fig. \ref{fig:suppl_archi}, the UNet network consists of down-block, mid-block, and up-block. Each of these blocks has 4, 1, and 4 subblocks respectively. Each of these subblocks typically constitutes two ResNet blocks and two inflated attention network blocks arranged as shown in Fig. \ref{fig:suppl_archi}. The only trainable components of the network belong to the inflated attention modules explained in the subsequent subsections. 

\vspace{2mm}
\noindent \textbf{Feature resolutions}: The VAE encoder $\mathcal{E}$ reduces the spatial dimensions from 768 to 96. The down-blocks further reduce the spatial dimensions to 24 while increasing the channel dimensions from 4 to 1280. The mid-blocks maintain the spatial dimensions and channel dimensions. The up-blocks increase the spatial dimensions to 96, while reducing the channel dimensions to 4. The VAE decoder then increases the spatial resolution back to 768. The CLIP text embedding and image embedding is a vector of size 768.

\vspace{2mm}
\noindent \textbf{UNet forward pass}: The inputs to the UNet network $\varepsilon_{\theta}(\cdot)$  are the latent noise from previous timestep $z_{t}$, the sinusoidal timestep embedding $t_{emb}$, an optional mask $\mathcal{M}$, the CLIP image embedding $\mathcal{J}$ and the CLIP text embedding $\mathcal{C}$. The latent noise $z_{t}$ is forwarded into layers of UNet network. At the end of a ResNet block, the hidden states are updated with the timestep embedding and the optional image embedding information based on the input mask --

\begin{itemize}
    \item \textbf{When an input mask is not provided, \emph{i.e.}, $\mathcal{M}=\mathcal{M}_{\phi}$:} In this situation, the hidden states are updated by adding the timestep embedding $t_{emb}$ and the image embedding $\mathcal{J}$ to all spatial locations of the hidden states.

    \item \textbf{When an input mask is provided, \emph{i.e.,} $\mathcal{M}\neq \mathcal{M}_{\phi}$:} In this situation, the regions that correspond to the background (\emph{i.e.,} where $\mathcal{M}=0$) are updated by adding the $t_{emb}$ and source image embedding $\mathcal{J}^{src}$. Similarly, for the regions corresponding to the foreground (\emph{i.e.,} where $\mathcal{M}=1$), the hidden states are updated by adding the timestep embedding $t_{emb}$ and the target image embedding $\mathcal{J}^{trg}$ as shown in Fig. \ref{fig:suppl_archi}. 
\end{itemize}
Once the hidden states are updated, they are passed into inflated attention blocks and the subsequent network layers. At the end of each denoising step of UNet, a latent fusion step is performed when an input mask is provided. In the next subsections, we explain the latent fusion method and the inflated model architecture.

\begin{figure*}[t]
    \centering
    \includegraphics[width=\linewidth]{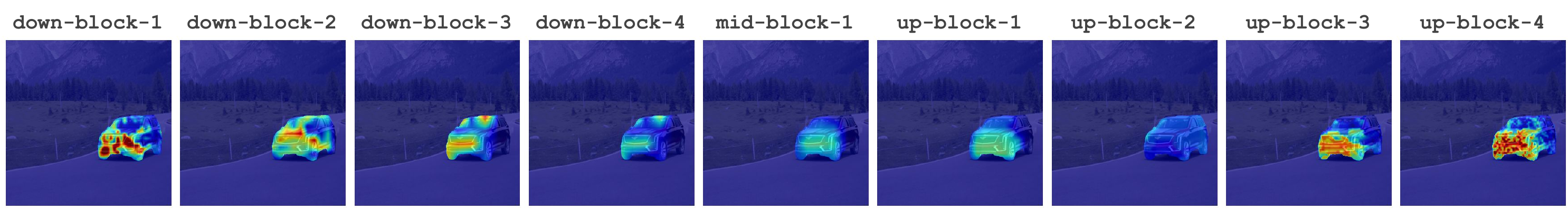}
    \caption{Correspondence Error (CE) maps computed using ground truth (source video correspondences) before correction across all blocks of UNet. We find that \texttt{Up-block-2} has the lowest CE.}
    \label{fig:suppl:cem}
\end{figure*}

\begin{figure*}[t]
    \centering
    \includegraphics[width=\linewidth]{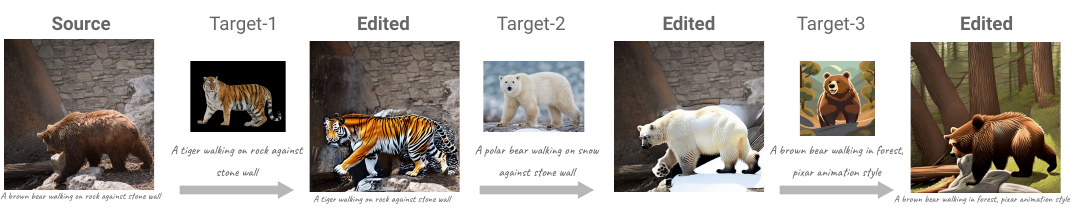}
    \caption{Zero-shot image editing results on the brown bear using \textit{InvEdit} mask. Background preservation is not used here. }
    \label{fig:suppl:zeroshot}
\end{figure*}

\vspace{2mm}
\noindent \textbf{Latent fusion}: Our latent fusion method follows from Make-A-Protagonist \cite{zhao2023makeaprotagonist}. The latent fusion step helps improve the quality of the rendered object in the edited video. First, UNet features are obtained using only the target image embedding $\mathcal{J}^{trg}$ with no mask input to UNet to obtain $\varepsilon_{\theta}(z,t,\mathcal{C}, \mathcal{J}^{trg}, \mathcal{M}_{\phi})$. Next, UNet features are obtained using source image embedding $\mathcal{J}^{src}$ and target image embedding $\mathcal{J}^{trg}$ along with a mask $\mathcal{M}$ to obtain $\varepsilon_{\theta}(z,t,\mathcal{C}, \{\mathcal{J}^{src}, \mathcal{J}^{trg}\}, \mathcal{M})$. Note here $\mathcal{J}^{src}$ is used for $\mathcal{M}=0$ region and $\mathcal{J}^{trg}$ is used for $\mathcal{M}=1$ region. These outputs are combined using the mask in the following manner:
    
\begin{align*}
    z_{t-1} &= \frac{1}{1+\mathcal{M}} \big(\mathcal{M} \odot \mathrm{DDIM}(\varepsilon_{\theta}(z_t,t,\mathcal{C}, \mathcal{J}^{trg}, \mathcal{M}_{\phi})) \\ &\quad ~~~ + \mathrm{DDIM}(\varepsilon_{\theta}(z,t,\mathcal{C}, \{\mathcal{J}^{src}, \mathcal{J}^{trg}\}, \mathcal{M}))\big)
\end{align*}

\begin{figure*}
    \centering
    \includegraphics[width=\linewidth]{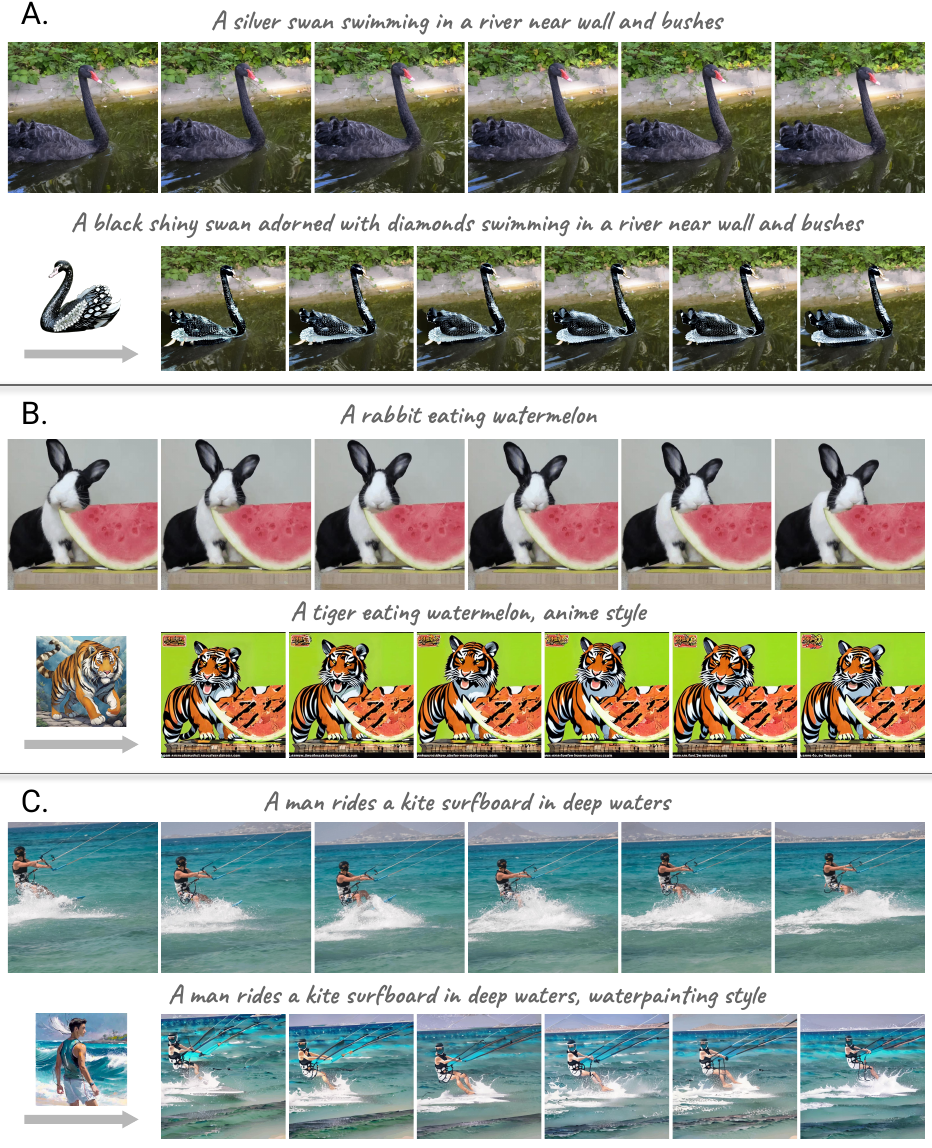}
    \caption{Additional results of \textit{GenVideo} on style editing of videos. Background preservation is not used in \textbf{B.} and \textbf{C.} since the entire video is being edited.}
    \label{fig:suppl:addn_style}
\end{figure*}

\noindent Note that when the background is allowed to be changed (like in Fig. \ref{fig:suppl:addn_style} and Fig. \ref{fig:suppl:zeroshot}), $\mathcal{J}^{src}$ is replaced with the CLIP image embedding of DALLE-2 prior obtained from the target text $\mathcal{P}^{trg}$. In all the other cases where the background is to be kept the same as the source, it is the CLIP image embedding of the source video frame, \emph{i.e.,} $\mathcal{J}^{src}$. More details can be found in \cite{zhao2023makeaprotagonist}.

\vspace{2mm}
\noindent \textbf{Inflated attention layers}: We follow the inflation strategy laid out by Tune-A-Video \cite{wu2022tuneavideo}. We expand the self-attention layers into spatio-temporal attention (ST-attn) layers by inputting the features from the first frames $g_{t,1}$ along with $g_{t,n-1}$ as shown in Fig. \ref{fig:suppl_archi} for computation of attention matrix. 
Here, $g$ denotes features of hidden states in the UNet. Cross-attention layers continue to accept the text tokens from prompt $\mathcal{C}$ along with  $g_{t,n}$. We additionally introduce temporal self-attention (T-attn) layers which are trained after permuting the temporal dimensions and spatial dimensions of the mini-batch. The only trainable weights in the entire pipeline are the query weights of ST-attn, query weights of Cross-attn, and all the weights in the T-attn as shown in Fig. \ref{fig:suppl_archi}.

\vspace{2mm}
\noindent \textbf{Additional details of training and inference pipeline}: During training, the source video is mapped into the VAE encoder's latent space. A random timestep is sampled and noise is added to the latents according to the forward diffusion process. The text embeddings of the source prompt and the image embeddings of a random source video frame are passed (as $\mathcal{C}$ and $\mathcal{J}$ respectively) into the UNet and the mask is $\mathcal{M}_{\phi}$. The reconstruction loss is imposed at the given timestep as shown in Fig. \ref{fig:method} of the main paper. The gradients are backpropagated using the AdamW optimizer to update the parameters of inflated attention modules described earlier. The inference pipeline consists of two stages - \textit{InvEdit} mask computation and the \textit{latent correction}. While computing the \textit{InvEdit} mask, the mask inputs to UNet are absent, \emph{i.e.,} $\mathcal{M}=\mathcal{M}_{\phi}$. After computing the \textit{InvEdit} mask $\mathcal{M}^{inv}$ it is passed into the UNet for mask guided inference and \textit{latent correction}, \emph{i.e.}, $\mathcal{M}=\mathcal{M}^{inv}$. See Algorithm 1 for inference pseudo-code.

\begin{figure*}
    \centering
    \includegraphics[width=\linewidth]{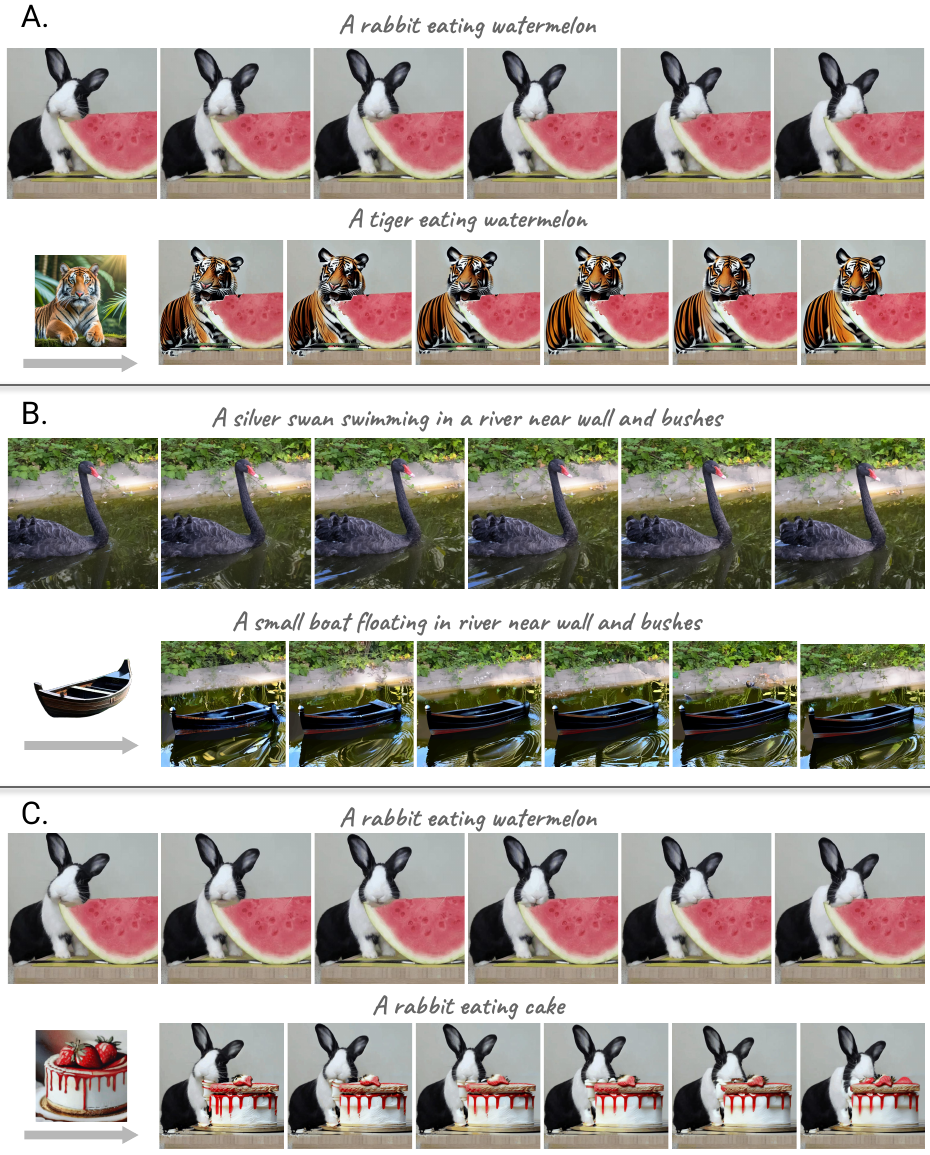}
    \caption{Additional results of \textit{GenVideo}. Our approach can do object edits when target-object has substantially different shape and size. }
    \label{fig:suppl:addn}
\end{figure*}

\algrenewcommand{\algorithmiccomment}[1]{\hfill $\triangleright$ \textit{#1}}

\begin{algorithm}
\caption{GenVideo Inference}
\begin{algorithmic}[1] 
\Require $\mathcal{V}^{src} := [I^{src}_{1:N}],  \mathcal{P}^{src}, \mathcal{P}^{trg}, I^{trg}$
\Require $ \varepsilon_{\theta}(\cdot) (\text{finetuned inflated UNet}), \mathcal{E}, \mathcal{D}, \mathrm{CLIP_{t}},  \mathrm{CLIP_{v}}$
\vspace{2mm}
\State \textbf{Set Hyperparameters:}
\State $\quad$ $T = 50$ \algorithmiccomment{DDIM timesteps}
\State $\quad$ $\alpha = 0.8$ \algorithmiccomment{Mask binarization threshold}
\State $\quad$ $w_{-1} = 0.1, w_{0}= 0.8, w_{-1}=0.1$ \algorithmiccomment{Inter-frame blending weights}
\vspace{2mm}

\State $\mathcal{C}^{src}, \mathcal{C}^{trg} = \mathrm{CLIP_{t}}(\mathcal{P}^{src}), \mathrm{CLIP_{t}}(\mathcal{P}^{trg})$
\vspace{1mm}
\State $ \mathcal{J}_{1:N}^{src}, \mathcal{J}^{trg} = \mathrm{CLIP_{v}}(I_{1:N}^{src}),  \mathrm{CLIP_{v}}(I^{trg})$
\vspace{1mm}
\State $ \mathcal{Z}^{src}_{T} := [z^{src}_{T,1}, \cdots, z^{src}_{T,N}] = \mathrm{DDIM}^{-1}(\mathcal{E}(\mathcal{V}^{src}))$ 
\vspace{1mm}
\State  ${z}^{trg}_{T, 1:N} = {z}^{src}_{T, 1:N}$
\vspace{2mm}

\For{$t \in [0.8\times T, T ]$} \algorithmiccomment{Compute the InvEdit mask}
\vspace{2mm}
    \State $t_{emb} = \mathrm{Emb}(t)$ \algorithmiccomment{sinusoidal timestep embedding}
    \State $ \textit{ }\varepsilon^{src}_{t, 1:N} = \varepsilon_\theta\left({z}^{src}_{t, 1:N}, t_{emb} , \mathcal{C}^{src}, \mathcal{J}^{src}_{1:N}, \mathcal{M}_{\phi} \right)$
    \State $\varepsilon^{trg}_{t, 1:N} = \varepsilon_\theta\left({z}^{trg}_{t, 1:N}, t_{emb} , \mathcal{C}^{trg}, \mathcal{J}^{trg}, \mathcal{M}_{\phi}\right)$
    \State $  \Delta\varepsilon_{t, 1:N} = \mathrm{abs}(\varepsilon^{src}_{t, 1:N} - \varepsilon^{trg}_{t, 1:N}) $
    \State $ z^{src}_{t-1, 1:N} = \mathrm{DDIM}(\varepsilon^{src}_{t,1:N})$
    \State $ z^{trg}_{t-1, 1:N} = \mathrm{DDIM}(\varepsilon^{trg}_{t,1:N})$

    \vspace{2mm}
\EndFor
    \State $ M_{1:N} = \mathrm{binarize}_{\alpha}(\mathop{\mathrm{mean}}_{t \in [0.8 \times T, T]}(\Delta\varepsilon_{t, 1:N}))$
    \State $\mathcal{M}^{inv} = M_{1:N} $
\vspace{2mm}
\For{$t=T, T-1, \cdots, 2$} \algorithmiccomment{Infer using InvEdit mask}
    \State 
    \State $[f^{t}_{1}, \cdots, f^{t}_{N}] \leftarrow \text{get } \texttt{Up-block-2} \text{ features }$
    \vspace{1mm}
    \State  $  \mathcal{N}^{t}_{i \pm}[p] = \mathrm{arg max}_{q} {d} (f^{t}_{i}[p], f^{t}_{i \pm 1}[q]), 1\leq i \leq N$  
    \vspace{1mm}
    \State $\mathcal{\hat{N}}^{t}_{i \pm} = \text{Upsample}(\mathcal{N}^{t}_{i \pm}) $ \algorithmiccomment{upsample to match the dim of $ \mathcal{Z}$ space}
    \State  $ o_{t, 1:N} = \varepsilon_{\theta}({z}^{src}_{t, 1:N}, t_{emb} , \mathcal{C}^{trg}, \mathcal{J}^{trg}, \mathcal{M}_{\phi})$  \algorithmiccomment{ UNet forward pass as in Sec.\ref{sec:arch_supp}}
    \vspace{2mm}
    \State  $ o'_{t, 1:N} = \varepsilon_{\theta}({z}^{src}_{t, 1:N}, t_{emb} , \mathcal{C}^{trg}, \{\mathcal{J}^{src}, \mathcal{J}^{trg}\}, \mathcal{M}^{inv})$  \algorithmiccomment{ UNet forward pass as in Sec.\ref{sec:arch_supp}}
    \vspace{2mm}
    \State    $ z_{t-1} = \frac{1}{1+\mathcal{M}^{inv}} \big(\mathcal{M}^{inv} \odot \mathrm{DDIM}(o_{t, 1:N})  + \mathrm{DDIM}(o'_{t, 1:N}) \big) $ \algorithmiccomment{ latent fusion as in Sec. \ref{sec:arch_supp}} 
    \vspace{2mm}
    \State  $        \Tilde{z}_{t-1, i}[p] = w_{-1} (M_{i} \odot  z_{t-1, (i-1)}[\mathcal{\hat{N}}^{t}_{i-} (p)])  + w_{0}( M_{i} \odot z_{t-1, i})  + w_{1}( M_{i} \odot z_{t-1, (i+1)}[\mathcal{\hat{N}}^{t}_{i+} (p)]) + (1 - M_{i}) \odot z_{t-1, i} , \text{ if } t \geq T-5 $ \algorithmiccomment{inter-frame latent correction }
    \vspace{1mm}
    \State Apply optional background preservation 
\EndFor
\vspace{2mm}
\State Output video frames = $\mathcal{D}(\Tilde{z}_{1, 1:N})$
\end{algorithmic}
\end{algorithm}

\section{Additional results}

\noindent \textbf{Selection of the UNet block for \emph{latent correction} field}: We compute the correspondence error (CE) map across all blocks of UNet as per Sec. \ref{sec:latentnoise} and find that the correspondence errors of \texttt{Up-block-2} are generally lower than other blocks as shown in Fig. \ref{fig:suppl:cem}. Across all experiments, we assign the feature with minimal Euclidean distance to the original feature as the corresponding feature. The correspondences obtained by computing RAFT optical flow on the source video serve as the ground truth since the object in the source video and the expected target object have the same shape. For computing the CE, we compare the correspondences obtained in the feature space with the ground truth from RAFT.

\vspace{2mm}
\noindent \textbf{Additional results of GenVideo}. In Fig. \ref{fig:suppl:zeroshot}, Fig. \ref{fig:suppl:addn} and Fig. \ref{fig:suppl:addn_style}, we present some additional results.
\begin{itemize}
    \item 

\vspace{1mm}
\textbf{Video object editing:}  In Fig. \ref{fig:suppl:addn}, we find that \textit{GenVideo} is able to accurately identify the region of interest to be modified. In Fig. \ref{fig:suppl:addn}{\color{BUred} A}, the \textit{InvEdit} mask accurately identified the region of edit and modified the region from the source \emph{rabbit} to the target \emph{tiger} while keeping the \emph{watermelon} intact. Similarly, in Fig. \ref{fig:suppl:addn}{\color{BUred} C}, the \emph{rabbit} was retained correctly and the region corresponding to the \emph{watermelon} was edited to \emph{cake} which has a different shape than the \emph{watermelon}. Thus, \emph{InvEdit} correctly handles the edits for varying shapes and sizes of objects. Results in Fig. \ref{fig:suppl:addn}{\color{BUred} B} demonstrate the editing of a \emph{silver swan} to a \emph{small wooden boat}. In this result, the \textit{InvEdit} mask helps in identifying regions that correspond to both \emph{swan} and expected \emph{boat} in order to edit the source video effectively even when they are of very different shapes and size here. 

\vspace{1mm}

\item \textbf{Style editing:} In Fig. \ref{fig:suppl:addn_style}, we present results of GenVideo for stylistic variation of the foreground object (in Fig. \ref{fig:suppl:addn_style}{\color{BUred} A}) and stylistic variations of the entire frames in the video (in Fig. \ref{fig:suppl:addn_style}{\color{BUred} B} and Fig. \ref{fig:suppl:addn_style}{\color{BUred} C}). When editing the entire frames in the video, we skip performing the background preservation.

\vspace{1mm}

\item \textbf{Zero-shot image editing:} In Fig. \ref{fig:suppl:zeroshot}, we show additional results on zero-shot image editing capabilities of our approach. 
\end{itemize}

{
    \small
    \bibliographystyle{ieeenat_fullname}
    \bibliography{main}
}

\end{document}